# Features and Machine Learning for Correlating and Classifying between Brain Areas and Dyslexia

Alex Frid
Laboratory of Clinical Neurophysiology
Rappaport Faculty of Medicine
Technion (Israel Institute of Technology)
Haifa, Israel
alex.frid@gmail.com

Larry M. Manevitz
Department of Computer Science
Ariel University, Ariel and University of Haifa
Haifa, Israel
manevitz@cs.haifa.ac.il

*Abstract*— **We develop a method that is based on processing gathered Event Related Potentials (ERP) signals and the use of machine learning technique for multivariate analysis (i.e. classification) that we apply in order to analyze the differences between Dyslexic and Skilled readers.**

**No human intervention is needed in the analysis process. This is the state of the art results for automatic identification of Dyslexic readers using a Lexical Decision Task. We use mathematical and machine learning based techniques to automatically discover novel complex features that (i) allow for reliable distinction between Dyslexic and Normal Control Skilled readers and (ii) to validate the assumption that the most of the differences between Dyslexic and Skilled readers located in the left hemisphere.**

**Interestingly, these tools also pointed to the fact that High Pass signals (typically considered as "noise" during ERP/EEG analyses) in fact contains significant relevant information.**

**Finally, the proposed scheme can be used for analysis of any ERP based studies.**

## I. Introduction

Dyslexia is a learning disability that impairs (-1.5s.d. and below[1]) a person's ability to decode words accurately and fluently [2]. This deficit can manifest itself as difficulties in phonological [3], orthographic [4] working memory [5], [6], brain systems asynchrony [4], [7], poor executive function skills [8] and/or rapid naming processing [6].

Clarifying and pinning down the neurological underpinning of dyslexia has been a major goal of research over the past twenty years. Amongst the theories proposed are ones focused on interhemispheric deficits, temporal asymmetries, phonological processing, and others. See [9]–[11] for a very comprehensive survey of the underlying neuro-physiological basis of Dyslexia.

The decoding of printed materials incorporates visual-orthographic, auditory-phonological and semantic processing which are based on the information that needs to be retrieved from the mental lexicon.

Furthermore, the reading activity requires the flow of relevant information from one brain area (the posterior lobes) to another (the frontal lobe). Whereas the posterior lobes are responsible for perception and physical processing of a stimulus, the frontal lobe provides meaning and motoric pronunciations to the stimulus [12]. In addition, word decoding relies on the transfer of information between brain hemispheres [12], [13]. The left hemisphere processes information in a sequential manner and specializes in linguistic processing. This hemisphere contains Wernicke's area, where the mental lexicon is stored, and Broca's area, which is responsible for language pronunciation. The right hemisphere for most right handed individuals processes information in a holistic way and specializes in the identification of visual patterns [12]. Reading as a linguistic activity requires work in both hemispheres [14] and diverse areas of the brain need to communicate in a timely fashion.

Experimental research has indicated that synchronization in time and level of activation between the visual and auditory brain systems at the perception level affects word decoding accuracy [3], [5], [7], [15]. It has also been proposed that speed of information processing in the brain has a crucial role in reading fluency for both dyslexic and typical readers [4], [7], [16].

Different studies have indicated that early visual, left inferior temporal and left superior temporal brain areas are activated in forming words patterns (for a review see [17]). In addition it is also clear that for the correct word decoding process to occur, these areas need to be activated separately as well as in an integrative fashion.

According to Breznitz [7], a gap in speed of processing (SOP) between the different brain entities activated in the word decoding process may prevent the precise synchronization of information necessary for an accurate process. This idea lies at the heart of the Asynchrony Theory, which suggests that the wider the SOP gap between the different brain entities; the more severe the word decoding failure will tend to be [7], [16]. (This outlook affected our choice of features somewhat; we discuss this aspect in a companion work.)

---

[1] A common criterion for diagnosing dyslexia is reading accuracy more than 1.5 SD (standard deviation) below the mean, which results in roughly 7% of the population being identified as dyslexic [1].

## II. SYSTEMS AND METHODS

### A. Experiment

The research included 32 native Hebrew-speaking children of grades 6-7 recruited from two schools. Of these, 17 were selected on the basis of previous diagnosis for dyslexia and 15 were verified as skilled readers. The reason for choosing this age group is the assumption that at this age the Dyslexia hasn't yet been fully compensated by other abilities, and thus it is suitable for searching its origins and true reflections.

Before conducting the EEG experiments, all participant taken a battery of tests in order to verify the general ability and their reading level.

All participants performed a Lexical Decision Task (LDT), during which event-related potentials (ERPs) were elicited by the presentation of words and pseudo-words. The stimuli were visually presented on the center of the computer screen and participants had to judge whether the letter strings seen were meaningful or meaningless. The participants were seated approximately 80 cm from an IBM-PC computer screen. Each subject was presented with a total of 96 moderate to high frequency words in the Hebrew language [18] and 96 pseudo-words created by substituting 1-2 letters in the real words. The task was presented to each subject in 4 blocks, with 24 trials in each of the blocks.

### B. Data Acquisition and Preprocessing

Electroencephalographic (EEG) signals were recorded using 64 scalp electrodes with a Biosemi system[2]. Scalp electrodes were referenced to an electrode on the chin and grounded to an electrode on the mastoid. Eye movements were monitored by a separate electrode placed below the left eye. The EEG data was sampled continuously at 2048Hz by using Biosemi ActiView software (Cortech Solutions, LLC), which is designed to display all ActiveTwo channels on screen and save all the data. During data collection, electrode impedance was kept below 5KΩ by prepping scalp areas with a mildly abrasive cleanser (NuPrep) and by using an electrolyte gel (Electro-gel). Signals were filtered at 0.1-100 Hz. The continuous EEG recordings were segmented off-line into 1.75-second epochs (448 samples, fs=256Hz). Each epoch contained 64 pre-stimulus samples (=250msec) for establishing baseline activity and 384 post-stimulus samples. Then, the EEG data was processed using Brain Vision Analyzer (Brain Products, GmbH). The data was filtered using a bandpass of 0.1 to 20 Hz, and baseline corrections were performed. Then the experimental conditions were averaged separately for each subject, after rejecting trials with incorrect behavioral responses. Grand average waveforms were afterwards computed for each experimental condition across all participants.

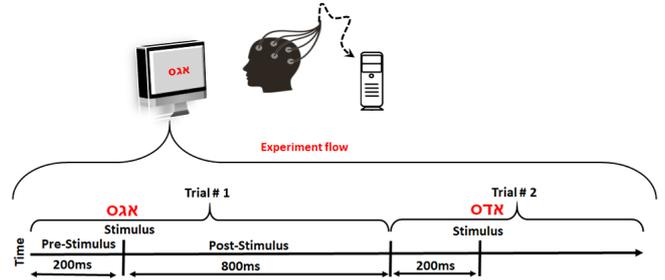

Fig. 1. Experiment flow.

### C. Analysis through Classification

Classification, or categorization, is the process of assigning a category (from a set of predefined categories) to a new observation (i.e. testing set), based on the training set of categorized data whose category is known. In this study, we refer to two categories, namely Dyslexic and Regular readers.

There are many advantages of using machine learning techniques (i.e. classification algorithms) compared to traditional statistical significance measures. One advantage is that by using classification, we can have a somewhat wider view on the problem through the way it treats new observations (the testing set). In our case these results could be also used later for diagnosis and screening purposes. Another advantage is in its ability to fuse sources of information in order to improve the success rate. In this work, we use this framework in a slightly different manner that allows us to not only accomplish the aforementioned goals, but also to present a novel method of ERP signal analysis in terms of brain regions of interest identification.

The classification scheme that was used in the present work was built up from the following steps: the preprocessing step, the feature extraction step, the classification step (which itself is broken up into training and testing phases), the decision making step and the feature selection step. The final methodologies that were used for each of the steps are described below and the resulting scheme can be seen in Fig. 1.

#### 1) Preprocessing Step

The preprocessing step, is the first step in the pipeline, and its goal is to prepare the data for the statistical analysis in terms of division of the signal into trials, artifacts removal (such as blinking, eye movements, loosed electrodes. baseline correction and filtering of the power line frequency). After the removal of the artifacts, the ERP signal is calculated by averaging the remaining trials.

#### 2) Feature Analysis and Extraction

In order to reduce the data dimensionality, which consisted of one second length (single trial) recorded from 64 electrodes sampled at 2048Hz, a set of descriptors or characteristics, (which in this context we refer to as "features") were extracted from each electrode lead. Besides dimensionality reduction, which leverages the capabilities of machine learning, we also wanted that these features should be chosen in a way that they are also meaningful (for humans) and potentially explanatory for understanding the dyslexic/skilled reader dichotomy.

---

[2] http://www.biosemi.com/

We carried out an extensive analysis of various characteristics of ERP signals, waveform shape related both in time and frequency domain. In the first stage, a Discrete Wavelet Transform [19] was carried out on the ERP signal. We then used this transformed signal to divide the ERP signal into two complementary parts: one based on information from the Low Pass (LP) data (i.e., low frequencies, also called the "Approximation" of the signal, or "A") and the High Pass (HP) data parts (i.e., high frequencies, also called the "Details" of the signal, or "D"). This transform is considered to be natural (as opposed to more traditional transforms such as the Fourier transform), since it has aspects that fit biological signals more closely [20].

From the LP part, a set of temporal features were extracted (mostly waveform shape related), and from the HP part a set of statistical and spectral features were extracted. This process of features analysis and extraction is described in more detail below:

*a) Division of the Signal into High and Low Frequency Components*

The first step in feature extraction was to divide the ERP signal into two. The first one consists of the low frequency components, and the second one consists of the high frequency components. This was done in order to extract different descriptors from each one of the parts. For this task, one-dimensional discrete Wavelet Decomposition [21] was used.

A wavelet [21], [22] is a wavelike signal with amplitude that starts at zero and gradually increases and then decreases back to zero. While a sinusoidal wave as is used by Fourier transforms carries on repeating itself for infinity, a wavelet exists only within a finite domain, and is zero-valued elsewhere. A wavelet transform involves convolving the signal against particular instances of the wavelet at various time scales and positions. Since we can model changes in frequency by changing the time scale, and model time changes by shifting the position of the wavelet, we can model both frequency and location of frequency. Thus the key advantage of the wavelet transform is in its temporal resolution (it can capture both the frequency and location information).

In this work, a Daubechies wavelet of order 4 (db4 wavelet) [23] was used, with the number (order) referring to the number of vanishing moments. The db4 was chosen due to its compromise between the smoothness and complexity, in addition to its asymmetry property. Since the waveform of the ERP signal is smooth and relatively simple the small orders of wavelets suffice to represent the signal.

Fig. 2 and Fig. 3 present the wavelet decomposition of the ERP signal from a regular and a dyslexic reader respectively. It can be seen that different temporal features such as maximum or minimum points can be easily detected. In addition, it can be seen that the HP parts of the signal for regular and dyslexic are quite different from each other in terms of periodicity and other frequency components. In common ERP analysis the HP parts are often called "noise" and removed from the signal as part of the signal preprocessing stage. In this work, however, we do include this part in the feature extraction (by extracting different statistical measures) and show that in fact this part contains valuable information and has a significant contribution to the overall classification rate.

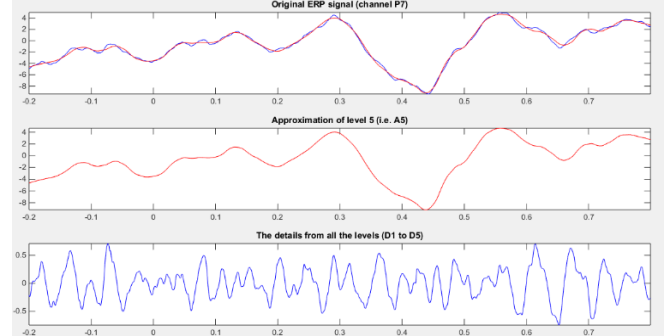

Fig. 2. The ERP pattern of a Regular reader from Electrode P7. First graph represents both the original ERP signal and its A5 approximation. The second graph represents only the A5 approximation. The third graph shows the detailed part from the wavelet approximation. The x-axis represents time in seconds, starting 200ms before the stimulus and ending 800ms after the stimulus (the stimulus itself occurred at time 0).

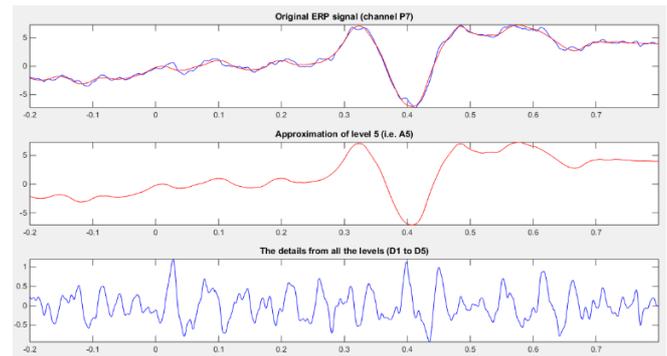

Fig. 3. The ERP pattern of a Dyslexic reader from Electrode P7. The first graph illustrates both the original ERP signal and its A5 approximation. The second chart presents only the A5 approximation. The third graph shows the detailed part from the wavelet approximation. The x-axis represents time in seconds, starting 200ms before the stimulus and ending 800ms after the stimulus (the stimulus itself occurred at time 0).

*3) Temporal Features Extracted from the LP part*

**Latency (L)** – The two times a maximal peak value appears in two predefined time segments. The two segments used in this study were 100ms-200ms and 200ms-400ms. This is designed to capture the existence and the strength of the relevant ERP signal components.

The first time frame is meant to catch the temporal occurrence of the Negative/Positive 150ms component that is usually associated with mapping of visual features (such as parts of letters ) onto higher level form representations (see [24] and [25]).

The second time frame is meant to catch the temporal occurrence of the P300 component that is usually elicited in the process of decision making, stimulus evaluation or categorization. Its latency varies with the difficulty of discriminating the target stimulus (and is thus expected to be different between the dyslexic and regular readers) [26], [27].

*Absolute Amplitude (AbsAmp)* – This is the maximal amplitude value in the signal. This feature stores the power of the strongest ERP component. A higher amplitude indicates that the summed epochs of event-related activity is more temporally consistent.

*Positive Area (Ap)* – This is the total amount of positive amplitudes in a selected time frame. The selected timeframes were between 100ms to 200ms, and 200ms to 400ms. These time frames were selected for the same reason as in the Latency feature above (i.e. to capture the P150 and P300 components), and to measure the overall energy of the signal. The total energy gives an indication about the strength of the response to a stimulus.

*Maximal Peak/Amplitude ratio (Mp)* – This is defined as the ratio between the maximal value of the amplitude in the defined time frame via its occurrence in time. Normalizing the peak magnitude by the time of its occurrence is a feature that maintains the information between a component's time occurrences (i.e. a late response will have a different value than a response with the same magnitude that occurred earlier in time).

*Overall Signal Energy (OSE)* – This is the overall energy of the signal. In discrete domain, the energy computed by:

$$E_x = \sum_{n=-\infty}^{n=\infty} |X(n)|^2 \quad (1)$$

where '$E$' is the total energy of a discrete signal X, and '$n$' is the sample index. This feature gives an indication of the overall activity of a specific ERP lead.

*Entropy* – This measures the uniformity of the histogram. It is defined as:

$$H(x) = -\sum_{i=1}^{N} P(x_i) \log_{10} P(x_i) \quad (2)$$

H(x) is the Entropy of the discrete signal, $x_i$ is a possible value that the signal can have, P($x_i$) is the probability of $x_i$ in the signal.

*4) Statistical and Spectral Features extracted from the HP part*

*Zero Crossing Rate (ZCR)* – the rate of the sign signals changes from the positive to negative or back.

$$ZCR = \frac{1}{N-1} \sum_{n=1}^{N-1} I\{x(n)x(n-1) < \} \quad (3)$$

Where '$N$' is the total signal length, '$I$' is an indicator function (i.e. equals to '1' when its argument is true and '0' otherwise).

*Mean (M), Standard Deviation (SD) and Skewness (Sk) ratios* – computed using the statistics of the time intervals between consecutive local derivative changes:

$$S = \frac{E(x - \mu(x))^3}{\sigma(x)^3} \quad (4)$$

where '$x$' is the signal value, $\mu$ is the mean of '$x$' and $\sigma$ is the standard deviation of '$x$'. Using this measure we can measure the asymmetry around the mean of the "noisy" part of the signal.

*5) Features related to structure of the frequencies*

To explain the point of these features, one needs to recall some basic facts about the Fourier transform and the frequency domain. Given an ERP signal, instead of looking at the information in the temporal domain; one can look at it in a frequency domain. The way of doing it is completely standard and often used the in digital signal analysis. The algorithm to compute the discrete Fourier Transform (DFT) is called Fast Fourier Transform (FFT) [28]

Accordingly, from this aspect, we found it useful to measure different spectral characteristics presented below.

*Spectral Flatness Measure (SFM)* – This estimates the flatness of a frequency spectrum (i.e. how flat the image is in spectral domain). SFM is calculated as the ratio between the geometric mean and the arithmetic mean of the frequencies:

$$F = \frac{\sqrt[N]{\prod_{N=0}^{N-1} X(n)}}{\sum_{N=0}^{N-1} X(n)} \quad (5)$$

Where "F" is the Spectral Flatness measure, and "X(n)" represents the magnitude value of bin "n" (in the frequency spectrum). This measure provides a way to quantify a stationary signal as opposed to being noise-like signal. This measure varies between 0 and 1; where being close to 0 indicates that the information is harmonical or stationary and is focused on few frequencies, while being close to 1 indicates that the signal is close to white noise.

*Spectral Roll-off (Sr)* – The spectral roll-off point defined as the N'th percentile of the power spectral distribution. The roll-off point is the frequency below which the N percent's of the magnitude distribution is concentrated. This feature can distinguish between high or low proportions of the energy contained in the high-frequency range of the spectrum, thus can be useful for detection of noisy ERP components.

**Power Spectral Density (PSD)**, Spectral Deformation, Spectral Width – Spectral deformation is defined as:

$$SD = \frac{\sqrt{M_2/M_0}}{M_1/M_0} \quad (6)$$

where $M_n$ is the nth spectral moment defined as

$$M_n = \sum_n P_i \cdot f_i^n \quad (7)$$

$M_2/M_1$ and $P_i$ is the power spectral density at frequency $f_i$.

*Spectral Centroid (Sc)* – This is a measure that indicates the spectral center of mass, is calculated as the weighted mean of the signal frequencies. It indicates where the "center of mass" of the spectrum is located. It is calculated as the weighted mean of the frequencies present in the signal, determined using a Fourier transform, with their magnitudes as the weights.

$$C = \frac{\sum_{n=0}^{N-1} f(n)x(n)}{\sum_{n=0}^{N-1} x(n)} \quad (8)$$

where x(n) represents the weighted frequency value, or magnitude, of bin number n, and f(n) represents the center frequency of that bin.

*Spectral Entropy (SE)* – the entropy (as shown in the equation below), is a measure of disorganization and it can be used to measure the "peakiness" of a distribution of the spectrum.

$$SE = \sum p_i \log_2(p_i) \quad (9)$$

The SE feature is calculated using the following steps:

i. Calculate the power spectrum of the signal using the Fast Fourier Transform (FFT).
ii. Calculate the Power Spectral Density (PSD) feature (as described above).
iii. Normalize the Power Spectral Density between [0-1], so that it can be treated as Probability Density Function (PDF) (marked as in the equation)
iv. Calculate the Entropy value by using the equation below.

*6) The Classification Procedure*

There are now quite a few different classification techniques that have been developed in machine learning. We considered the advantages and disadvantages of several such as Decision Trees [29], Principal Component Analysis (PCA) [30], basic neural networks [31], Support Vector Machines (SVM) [32], [33], Naïve Bayes and others.

We rejected methodologies such as Naïve Bayes which require much more data than we had available, and after preliminary experimentation related methods such as Decision Trees [29]. The other methods listed above were all implemented, but the best and most robust of the methods was Support Vector Machines.

*7) Detection of Regions of Interest*

We are interested in understanding the areas of the brain that are involved in the reading process and their correlation with dyslexia. That is, in addition to developing the ability of correct and accurate classification rate, we also wish to understand the root cause of the problem i.e. to understand what parts of the brain are involved in the process and exactly what features found to be the most descriptive. We relate this to the classification process by seeking features that hold most of the information needed for correct classification. Those areas that holds the most important information can be called Regions of Interest (ROIs). Thus, in order to find the ROIs, a features selection method is applied.

In other words, our goal is to reduce the dimension of the data by finding a small set of important features which can still provide good classification results.

Feature selection algorithms can be divided into two major categories: (i) filter based methods and (ii) wrapper based methods. The filter based methods rely on general statistical characteristics of the data in order to evaluate and select the feature subsets (i.e. without involving the learning algorithm itself). On the other hand, the wrapper methods search for features that fit well to the learning algorithm that is used. One judges the quality of a feature by how much it contributes to a run of the learning algorithm. (As a consequence they can be significantly slower than filter methods if the learning algorithm takes a long time to run.) For more on this dichotomy see [34].

In this work a ReliefF method [35] was used for the task of feature selection.

**Feature Selection using ReliefF algorithm**

In order to automatize the process of ERP analysis a global automated feature extraction process is carried out. The goal here is two-fold: not only to have features that can successfully separate (i.e. classify) between the groups, but also be able to indicate a brain-wise localization of the origin of the problem. In addition, the features should be such that they are descriptive (i.e features that can be interpretable directly by humans). Having selected such features we can extract them from all the signals.

In other words, the goal is not only to have the ability to successfully classify (separate) between the groups but also to indicate the origin, or the root cause of the problem (done by the feature selection). (We can call this second ability "Backtracking".)

The advantage of the ReliefF algorithm is in being a bit from both the filter and wrapper approaches. From one hand it uses the K-nearest neighbor algorithm inside, and thus might be considered as part of the wrapper methods, and from the other hand its running speed is close to filter method algorithms. Another important advantage of this algorithm lies in it being a "non-parametric feature weighting algorithm" (i.e. it makes no assumption about the distribution of the data).

The *ReliefF* algorithm is a generalization of the *Relief* algorithm that is presented in [36]. The basic idea of the algorithm is to measure the relevance of features in the neighborhoods around the data samples. For each data sample, the basic Relief method finds the nearest sample in feature space of the same category, called the "hit" sample, and the nearest sample of the other category, called the "miss" sample. Where the overall relevance of feature is the sum of its relevance for all target samples.

An extension of this algorithm (i.e. the ReliefF algorithm) simply uses a set of nearby "hits" and a set of nearby "misses" for every target sample and averages their distances in order to be more robust to noise. As shown in the following equation (Eq. 10):

$$W_f \mathrel{+}= \frac{\sum_{m \in misses} |f_{target} - f_{miss}| - \sum_{m \in hit} |f_{target} - f_{hit}|}{|misses| \cdot |hits| \cdot Range(f)} \quad (10)$$

In general, the algorithm assign relevance to features based on their ability to disambiguate similar samples, where similarity is defined by proximity in feature space. Relevant features accumulate high positive weights, while irrelevant features retain near-zero weights.

Due to its low complexity, this algorithm is ideal for domains where the number of features is much larger than the number of training samples. As it is in our case, where the same set of selected features extracted from all the electrodes, where 27 handpicked features were automatically extracted from each of the 64 electrodes (i.e. feature vector of 1728 features). The ReliefF was activated with K=10 (i.e. 10 nearest neighbors), the x-axis represents the feature number and the y-axis its importance (i.e. the weight). Interestingly, it can be seen that a large amount of features has a negative effect on the classification.

The overall scheme of the algorithm is summarized in Fig. 4. Starting from the data acquisition stage, the preprocessing and the ERP generation stages, followed by the feature extraction, feature selection (using ReliefF algorithm, as described above) and finally the analysis of region of interest.

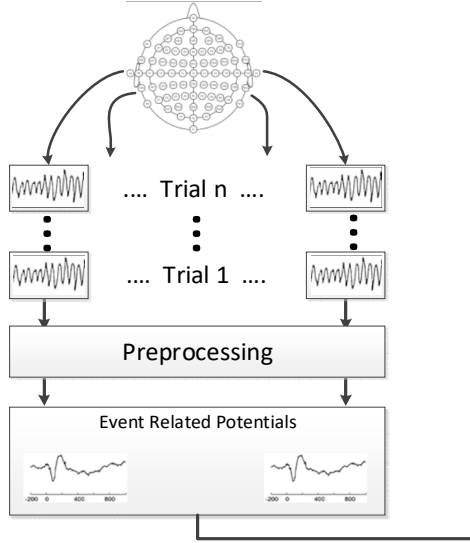
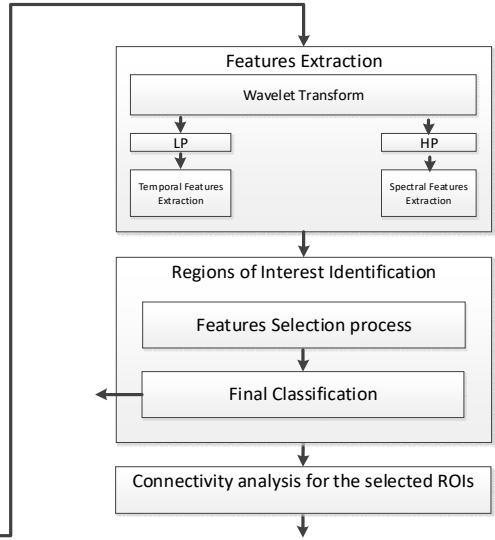

Fig. 4. Overview of the features extraction, features selection and classification processes

III. RESULTS

In order to summarize the ability of the method, we report three types of results. The first result, presented as a full confusion matrix in Table 1 is the classification results using the best 60 features selected by the feature weighting algorithm.

TABLE I. CONFUSION MATRIX FOR THE CLASSIFICATION RESULTS USING ONLY THE BEST 60 FEATURES

|  | Regular | Dyslexic |
|---|---|---|
| Regular | **79%±1.9%** | 21% |
| Dyslexic | 22% | **78%±2.3%** |

The second result, presented in Table 2 is the classification results using only the 10 best features. It can be seen that this subset of 10 features carries most of the information needed in order to discriminate between the Dyslexic and Regular readers.

TABLE II. CONFUSION MATRIX FOR THE CLASSIFICATION RESULTS USING THE BEST 10 FEATURES

|  | Regular | Dyslexic |
|---|---|---|
| Regular | **70%±2.1%** | 30% |
| Dyslexic | 29% | **69%±1.6%** |

Due to the success of the second result, we present another type of analysis that is depicted in Fig 5. (The selected areas are marked by the common convention.) The idea is that we cannot relate the localization to a single electrode due to the fact that it may vary a bit from head to head, and thus, even if a single electrode is chosen, still the entire area in which this electrode is related is chosen as the area of interest. Under this method, the number of electrodes from the same area that have a major contribution over all classification indicates the relative importance of this area. The final results of the feature selection described in Fig 5 are four areas containing the electrodes that are important to the discrimination between the dyslexic and regular readers. In addition, it can be seen that the left anterior part is of higher importance (most of the electrodes that contributes for the overall classification are located there). Finally, another conclusion that can be reflected from this, is that the most of the differences between the dyslexic and regular readers are located in the left hemisphere.

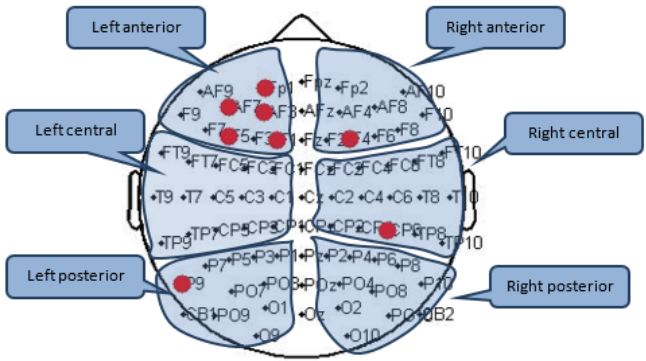

Fig 5- The results after the feature selection procedure. The red electrodes are the selected electrodes which contain most of the difference between Dyslexic and Regular readers. It can be seen how the differences concentrated in very specific areas.

We wish to point out that the pipeline described above and the subsequent labeled image in Fig. 5 is general and can be applied in any ERP study of a cognitive task.

TABLE III.   THE MOST IMPORTANT FEATURES EXTRACTED BY THE FEATURE SELECTION ALGORITHM

| Feature name and short description | Extracted from |
|---|---|
| Latency (L) of the p300 peak | LP |
| Absolute Amplitude (AbsAmp) of the signal | LP |
| Max.Peak Time/Max.Peak Amplitude ratio (MP) | LP |
| Skewness (Sk) of the zero crossing time intervals | HP |
| Mean (M) of the zero crossing time intervals | HP |
| Spectral Reoll-off (Sr) point (70% cutoff) | HP |
| Signal Energy (OSE) | HP |

## IV. SUMMARY AND DISCUSSION

In this work we investigated if advanced signal analysis combined with machine learning classification techniques are sufficient to find biological markers for Dyslexia.

To explore this work, we instituted the following techniques:

(i) We have shown that the data contains enough information in order to discriminate between the Dyslexic and the Regular readers. We have shown success of close to 80% in discrimination in this task. This analysis is performed on the ERP signal (i.e. on an average of EEG trials). This is performed by using automatic feature extraction techniques applied on all the electrodes and followed by a feature selection technique.

(ii) This methodology allowed us to go back and find the most important brain areas for the task (as seen in Fig. 5). These results in fact do correspond with current theories regarding the role of the left hemisphere in reading process in developmental Dyslexia [37], [38].

(iii) Most of the information needed for the discrimination is in fact concentrated in parts of the signal that is traditionally considered as noise and in fact typically removed before analysis. These features are listed in Table 3, since some of the features selected by more than one electrode; the number of the features in the table is smaller[3].

Thus, the main results that were established in this paper are:

Using machine learning tools, a small number of features suggested by signal analysis techniques are sufficient to accurately classify dyslexic / skilled readers from EEG data summed up to an ERP signal.

A method based on feature weighting for "Regions of Interest" detection was proposed.

Parts of the signal considered as noise by traditional analysis techniques are actually contains valuable information.

ACKNOWLEDGMENT

This work was partially supported by a grant for computational equipment by the Caesarea Rothschild Institute and by a Hardware Grant by NVIDIA Corporation to the Neurocomputation Laboratory.

---

[3] Having ascertained that the signal contains enough information, in other work we then investigated if the root of the problem lies in the synchrony between different brain modalities. This was done by analysis that is applied on the raw EEG signal (i.e. not averaged) and by looking on global synchronizations.